\documentclass{article}
\usepackage[preprint]{colm2026_conference}
\usepackage[T1]{fontenc}
\usepackage{microtype}
\usepackage{hyperref}
\usepackage{url}
\usepackage{booktabs}
\usepackage{graphicx}
\usepackage{subfigure}
\usepackage{listings}
\usepackage{xcolor}
\usepackage{tikz}
\usetikzlibrary{shapes,arrows,positioning,decorations.pathreplacing}
\usepackage{amsmath}
\usepackage{amssymb}
\usepackage{mathtools}
\usepackage{amsthm}
\usepackage{lineno}
\usepackage[most]{tcolorbox}
\tcbuselibrary{skins,breakable}

\definecolor{PromptBlue}{HTML}{3B4BE3} 
\newtcolorbox{promptwindow}[1]{%
  enhanced,
  breakable,
  colback=PromptBlue!6,
  colframe=PromptBlue,
  boxrule=1pt,
  arc=14pt,
  left=14pt,right=14pt,top=14pt,bottom=14pt,
  fonttitle=\bfseries,
  coltitle=white,
  colbacktitle=PromptBlue,
  attach boxed title to top left={xshift=10pt,yshift=-2mm},
  boxed title style={
    sharp corners,
    arc=10pt,
    boxrule=0pt,
    left=10pt,right=10pt,top=6pt,bottom=6pt
  },
  title=#1
}

\newtcolorbox{promptwindowwhite}[1]{%
  enhanced,
  breakable,
  colback=white,
  colframe=PromptBlue,
  boxrule=1pt,
  arc=14pt,
  left=14pt,right=14pt,top=14pt,bottom=14pt,
  fonttitle=\bfseries,
  coltitle=white,
  colbacktitle=PromptBlue,
  attach boxed title to top left={xshift=10pt,yshift=-2mm},
  boxed title style={
    sharp corners,
    arc=10pt,
    boxrule=0pt,
    left=10pt,right=10pt,top=6pt,bottom=6pt
  },
  title=#1
}

\definecolor{darkblue}{rgb}{0, 0, 0.5}
\hypersetup{colorlinks=true, citecolor=darkblue, linkcolor=darkblue, urlcolor=darkblue}

\definecolor{codegreen}{rgb}{0,0.6,0}
\definecolor{codegray}{rgb}{0.5,0.5,0.5}
\definecolor{codepurple}{rgb}{0.58,0,0.82}
\definecolor{backcolour}{rgb}{0.95,0.95,0.92}

\lstdefinestyle{mystyle}{
    backgroundcolor=\color{backcolour},
    commentstyle=\color{codegreen},
    keywordstyle=\color{blue},
    numberstyle=\tiny\color{codegray},
    stringstyle=\color{codepurple},
    basicstyle=\ttfamily\small,
    breakatwhitespace=true,
    breaklines=true,
    captionpos=b,
    keepspaces=true,
    numbers=none,
    showspaces=false,
    showstringspaces=false,
    showtabs=false,
    tabsize=2
}
\theoremstyle{plain}

\theoremstyle{definition}

\theoremstyle{remark}

\title{FormalRewardBench: A Benchmark for Formal Theorem Proving Reward Models}

\author{Zeynel A. Ulusan$^{1,2}$ \quad Burak S. Akbudak$^{3}$\thanks{Equal contribution.} \quad Can S. Erer$^{3}$\footnotemark[1] \quad Gözde Gül \c{S}ahin$^{1,4}$ \\
$^{1}$Koç University, Department of Computer Science and Engineering\\
$^{2}$Codeway Studios \\
$^{3}$Boğaziçi University, Department of Computer Engineering  \\
$^{4}$Friedrich-Alexander-Universität Erlangen-Nürnberg, Intelligent Language Systems\\
\texttt{ \url{https://gglab-ku.github.io/}}
}

\begin{document}

\maketitle

\begin{abstract}
  Recent neural theorem provers use reinforcement learning with verifiable rewards (RLVR), where proof assistants provide binary correctness signals. While verifiable rewards are cheap and scalable without reward hacking issues, they suffer from sparse credit assignment: models receive no learning signal from difficult problems where partial progress goes unrewarded. This motivates learned reward models that can evaluate proof quality beyond binary verification. However, comparing reward models is challenging since it typically requires expensive RL training ablations. To address this, we introduce \textbf{FormalRewardBench}, the first benchmark for evaluating reward models in formal theorem proving with Lean 4. Our benchmark consists of 250 preference pairs where correct proofs are paired with incorrect variants generated through five expert curated error injection strategies: forced mistakes, minimal single-point variations, verbose incorrect proofs, natural language justification, and Python code injection. We evaluate frontier LLMs (e.g., Claude Opus 4.5), judge LLMs (e.g., CompassJudger-1-14B), general-purpose LLMs (e.g., Qwen2.5-72B-Instruct), and specialized theorem proving models (e.g., DeepSeek-Prover-V2-7B).
  Our results reveal that frontier LLMs achieve the highest performance (59.8\%) while specialized theorem provers perform the worst (24.4\%), suggesting that theorem proving ability does not transfer to proof evaluation. We provide further insights on various error injection mechanisms, highlighting the challenging nature of most injection mechanisms. We release \textbf{FormalRewardBench} publicly to encourage more research on developing reward models in formal mathematics.  
\end{abstract}

\section{Introduction}

Formal theorem proving provides machine-verifiable guarantees for mathematical claims, with applications in software verification and formalizing research mathematics. Large language models have achieved remarkable results on this task, including gold-medal-level performance at the International Mathematical Olympiad~\citep{alphaproof2024} and solving challenging problems from Putnam competitions. Recent systems like DeepSeek-Prover~\citep{deepseek-prover-v15, deepseek-prover-v2} and G\"{o}del-Prover~\citep{godel-prover} leverage neural language models to generate formal proofs in proof assistants such as Lean~4, achieving strong results on challenging benchmarks like MiniF2F~\citep{minif2f} and PutnamBench~\citep{putnam-bench}.

Much of this progress comes from reinforcement learning with verifiable rewards (RLVR), where the proof assistant's type checker provides binary correctness signals. Unlike learned reward models that can suffer from overoptimization and reward hacking, verifiable rewards are cheap, scalable, and perfectly accurate. However, this binary feedback creates a fundamental limitation: \textbf{sparse credit assignment}. A proof attempt that makes substantial progress but fails at the final step receives the same zero reward as a completely wrong approach. Models cannot learn from difficult problems where they are on the right track but cannot complete the proof. This is a well-known challenge in reinforcement learning, where sparse rewards limit effective credit assignment~\citep{sutton2018reinforcement}.This sparsity problem motivates learned reward models that can evaluate proof quality beyond binary verification. While methods such as DPO~\citep{dpo} and GRPO~\citep{deepseek-prover-v15} bypass explicit reward models, theorem proving benefits from richer reward signals for both training and inference-time proof selection~\citep{proofverifier2025, snell2024scaling}. However, comparing such reward models typically requires expensive RL training ablations.

\begin{figure}[t]
    \centering
    \includegraphics[width=0.75\columnwidth]{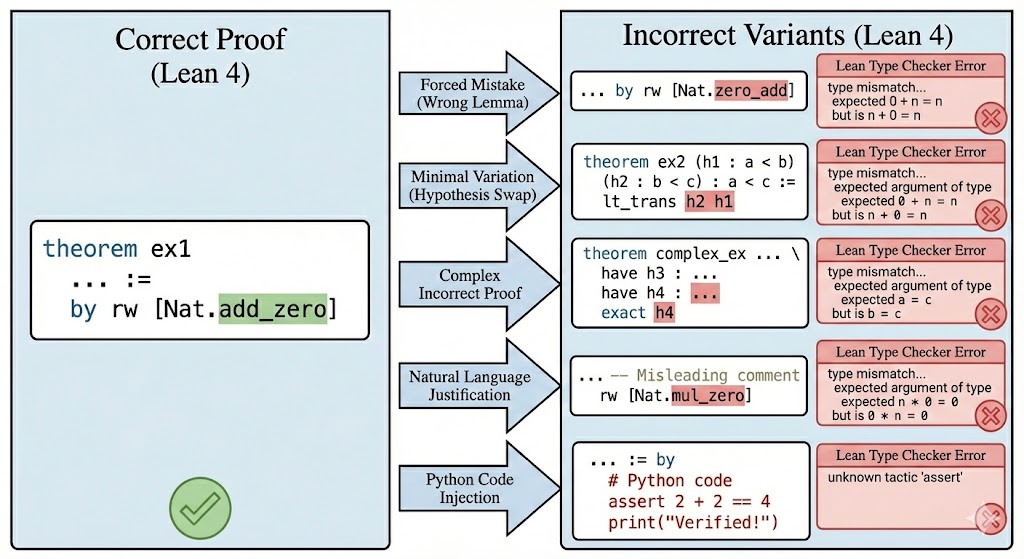}
    \caption{Overview of the error injection pipeline employed in FormalRewardBench. A formally verified correct Lean 4 proof (left) is transformed into incorrect variants (right) using five distinct strategies: Forced Mistakes, Minimal Single-Point Variations, Verbose Incorrect Proofs, Natural Language Justification, and Python Code Injection. While the generated variants remain syntactically valid and appear plausible to language models, they fail formal verification (type checking) as shown by the error messages.}
    \label{fig:error-example}
\end{figure}
We address this by introducing \textbf{FormalRewardBench}, a benchmark for directly evaluating reward models on formal theorem proving. Similar to how RewardBench~\citep{rewardbench} enables reward model comparison for RLHF without training runs, FormalRewardBench enables comparison for formal reasoning without RL ablations. Our benchmark consists of preference pairs where correct proofs are paired with incorrect variants, allowing direct measurement of whether models can distinguish valid from invalid proofs.
We build our benchmark from MiniF2F \citep{minif2f}, a dataset of 488 olympiad-level mathematical problems formalized in Lean 4, covering algebra, number theory, and combinatorics from competitions like AMC, AIME, and IMO. We generate incorrect proof variants through five error injection strategies designed by the authors, who have formal training in mathematics, and implemented via carefully crafted LLM prompts (see Figure 1 and §3.1.2). These strategies target different dimensions of proof correctness: (1) \textit{minimal single-point variations} that make small but semantically impactful edits, (2) \textit{natural language justification} that augments incorrect proofs with misleading explanatory comments, (3) \textit{Python code injection} that replaces formal proof steps with executable Python code, (4) \textit{forced LLM mistakes} such as applying the wrong proof rule or referencing the wrong assumption, and (5) \textit{verbose incorrect proofs} with lengthy but fundamentally flawed reasoning.
We evaluate four categories of models: frontier LLMs (e.g., Claude Opus 4.5, GPT-5.2), judge LLMs trained for preference evaluation (e.g., Skywork-Critic-Llama-3.1-70B, CompassJudger-1-14B-Instruct), general-purpose LLMs trained on math and code domains (e.g., Qwen2.5-72B-Instruct, DeepSeek-Coder-V2), and theorem proving specialized models (e.g., DeepSeek-Prover-V2-7B, G\"{o}del-Prover-V2-8B), ranging from 7B to 72B parameters. Our results reveal that frontier LLMs achieve the highest performance (70.1\% pointwise), judge models substantially outperform specialized theorem provers among open-weight models (52.8\% vs 12.8\%), and error categories vary widely in difficulty.

Our contributions are threefold: (1) \textbf{FormalRewardBench}, the first benchmark for evaluating reward models in formal theorem proving, consisting of 250 preference pairs; (2) \textbf{five expert-curated error injection strategies} that form a reusable methodology for generating realistic incorrect formal proofs, applicable to other proof assistants, training data augmentation, or curriculum design for theorem provers; and (3) \textbf{comprehensive evaluation} revealing that general reward modeling capabilities transfer more effectively than domain-specific training. Developing effective reward models for formal theorem proving could accelerate mathematical discovery and verification. Automated theorem provers with nuanced feedback could assist mathematicians in formalizing complex proofs and verify software and hardware systems more efficiently. However, overreliance on imperfect reward models carries risks: our work highlights current models' limitations, emphasizing the continued necessity of formal verification as the ultimate arbiter of correctness. FormalRewardBench is publicly available at \href{https://github.com/GGLAB-KU/formal_rewardbench}{FormalRewardbench}.

\section{Related Work}

\paragraph{Reward Models for Reasoning.} RLVR uses external verifiers (proof assistants, code executors, symbolic solvers) to provide binary correctness signals~\citep{reinforcement-learning-from-human-feedback}. While scalable, this binary feedback provides sparse reward signals that limit learning from partial progress~\citep{deepseek-prover-v15}. Two approaches provide denser signals: LLM-as-Judge prompts models to compare responses directly~\citep{zheng2023judging}, and Generative Reward Models (GenRMs) produce natural language explanations before judgment~\citep{genrm, genprm}. While these approaches have been extensively studied for informal reasoning tasks, their effectiveness in formal theorem proving remains unexplored. FormalRewardBench addresses this gap by providing the first benchmark to evaluate these reward modeling approaches on formal proofs.

\paragraph{Reward Model Benchmarks.}
RewardBench~\citep{rewardbench} and its successor RewardBench 2~\citep{malik2026rewardeval} evaluate reward models across chat, safety, and reasoning categories. ProcessBench~\citep{processbench} and PRMBench~\citep{prmbench} evaluate process reward models for step-level mathematical reasoning. However, all operate on informal natural language outputs. Formal theorem proving requires syntactic precision, type correctness, and logical soundness~\citep{lean4}, where proofs that appear reasonable informally may contain subtle errors only apparent during formal verification. No existing benchmark evaluates reward models for formal proofs.

\paragraph{Neural Theorem Proving.}
DeepSeek-Prover~\citep{deepseek-prover-v15, deepseek-prover-v2}, G\"{o}del-Prover~\citep{godel-prover}, and Kimina-Prover~\citep{kimina-prover} achieve strong results on MiniF2F~\citep{minif2f} (488 olympiad problems), ProofNet~\citep{proofnet} (undergraduate mathematics), and PutnamBench~\citep{putnam-bench} (1697 competition problems). These benchmarks use pass@k metrics, capturing success but not proof quality or error severity. FormalRewardBench complements them by evaluating the ability to distinguish correct from incorrect proofs.

\section{Methodology}

Our benchmark targets Lean 4, a proof assistant based on dependent type theory where proofs are constructed via tactics and verified by a type checker (see Appendix~\ref{app:background} and Appendix~\ref{app:reward-models}  for details on Lean 4 and reward modeling formulations). We create a benchmark for evaluating reward models on formal theorem proving that goes beyond binary outcome evaluation. The core design philosophy is controlled difficulty through synthetic error injection. We begin with formally verified correct proofs and systematically generate incorrect variants that satisfy three criteria:
\begin{itemize}
    \item \textbf{Syntactic validity:} The generated incorrect proof must parse as valid Lean code.
    \item \textbf{Semantic plausibility:} Errors should not be trivially obvious (e.g., undefined variables, basic syntax errors).
    \item \textbf{Realistic error patterns:} Mistakes should reflect actual failures of language models on the task.
\end{itemize}

This approach allows us to create a challenging benchmark that measures genuine formal proof understanding rather than surface-level pattern matching.

\subsection{Dataset Construction}

We build our benchmark from MiniF2F~\citep{minif2f}, a dataset of 488 olympiad-level mathematical problems formalized in Lean 4. Problems span algebra, number theory, and combinatorics from competitions including AMC, AIME, and IMO. Data sources vary by strategy: For S1 (Minimal Change), we use correct proofs from DeepSeek-Prover-V2-671B and prompt an LLM to introduce minimal errors. For S2 (NL Justification), we randomly sample incorrect proofs from DeepSeek and G\"{o}del model families, then augment them with misleading comments. For S3, S4, and S5 we prompt LLMs to generate incorrect proofs directly from theorem statements.

\begin{figure}[th]
    \centering
    \includegraphics[width=\columnwidth]{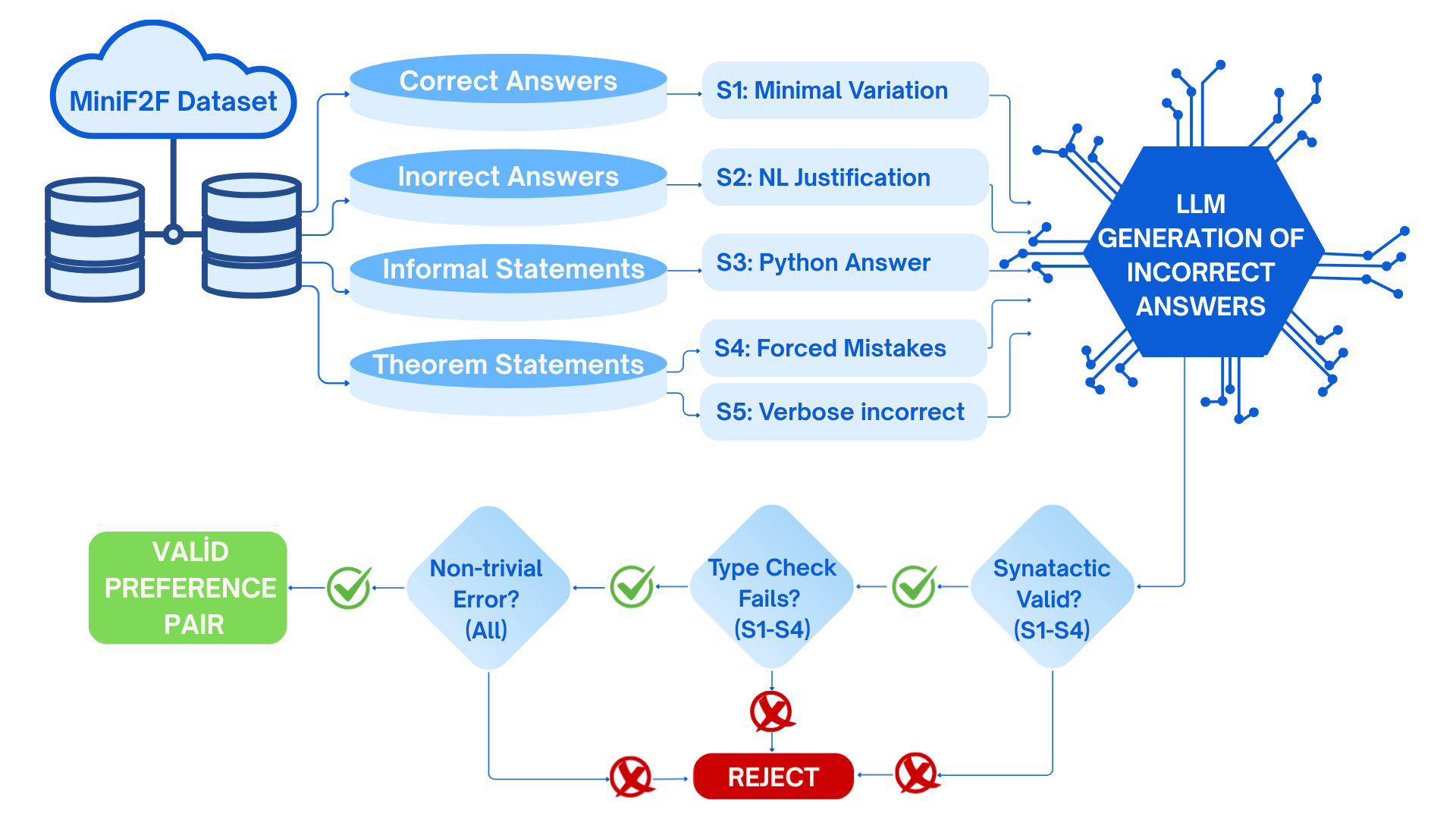}
    \caption{FormalRewardBench generation pipeline. Top: Data flows from MiniF2F through five strategies to LLM generation. Each strategy uses different inputs: correct proofs (S1), incorrect proofs (S2), informal statements (S3), or formal statements (S4, S5). Bottom: Quality control filters. S1--S4 require syntactic validity and type check failure. All strategies filter trivial errors. S3 bypasses Lean validation as Python code.}
    \label{fig:pipeline}
\end{figure}

\subsubsection{Generation Pipeline and Quality Control}
\label{ssec:qualityPipeline}

Figure~\ref{fig:pipeline} illustrates our data generation pipeline. We source correct proofs from DeepSeek-Prover-V2-671B~\citep{deepseek-prover-v2}, the strongest open-source theorem prover at the time of data collection, which provides verified solutions for a substantial subset of MiniF2F problems. Incorrect proofs are sampled from the DeepSeek-Prover and G\"{o}del-Prover~\citep{godel-prover} model families, as their failed proof attempts represent realistic LLM failure modes on these problems.

Each error injection strategy uses different source data: S1 (Minimal Variation) takes correct proofs and introduces minimal errors; S2 (NL Justification) takes incorrect proofs and adds misleading comments; S3 (Python Answer) takes informal statements and generates Python code; S4 (Forced Mistakes) and S5 (verbose incorrect) take formal theorem statements and generate incorrect proofs directly.

For error generation, we use Claude Opus 4.5 as the prompting model across all five strategies. We chose a frontier general-purpose LLM rather than a theorem proving specialist to avoid biasing the generated errors toward any single prover's failure distribution, ensuring broader coverage of possible error patterns. For each theorem and strategy, we generate 3 candidates. Generated proofs pass through three quality filters: (1) \textit{syntactic validation} (S1--S4) verifies that the proof parses in Lean 4; (2) \textit{semantic validation} (S1--S4) confirms that the proof fails Lean's type checker; and (3) a \textit{triviality filter} rejects proofs with superficial errors (e.g., ``unexpected token'') while accepting semantic errors (e.g., ``type mismatch''). S3 (Python Answer) bypasses Lean validation since Python code does not parse as Lean. We randomly sample 50 validated pairs from each strategy, yielding 250 preference pairs total. The choice of 50 pairs per strategy is justified by our stability analysis (Appendix~\ref{app:stability}), which shows that model accuracy stabilizes around this threshold.

\textbf{Quality Assurance.}
Our benchmark employs two-layer quality assurance. First, \textit{automatic verification}: for S1--S4, every correct proof passes Lean's type checker and every incorrect proof fails it, providing objective, deterministic correctness labels unlike informal benchmarks that rely on subjective human judgment. Second, \textit{manual inspection}: the authors reviewed a random sample of 50 pairs (10 per strategy) to verify that (i) incorrect proofs are syntactically plausible, (ii) errors are non-trivial and require genuine proof understanding to detect, and (iii) error patterns are realistic.

\vspace{-0.25em}

\subsubsection{Error Injection Strategies}
\label{ssec:errorTypes}
 Since we source data from actual LLM outputs and use LLMs to generate error variants, our benchmark naturally captures realistic LLM failure modes. All strategies are implemented using carefully designed prompts; details in Appendix~\ref{app:prompts}. We provide concrete examples for all five strategies in Appendix~\ref{app:error-examples}.

\textbf{Strategy 1: Minimal Single-Point Variations.}
This strategy creates incorrect proofs by making minimal, surgical single-point modifications with maximal semantic impact. Valid variations include changing a single variable, introducing type mismatches (e.g., replacing \texttt{Int} with \texttt{Nat}), or using the wrong hypothesis among similar ones (e.g., swapping \texttt{h1} and \texttt{h2}). Identifying which modifications are semantically meaningful rather than causing trivial syntax errors requires strong reasoning, code comprehension, and instruction-following capabilities, which motivates the use of frontier LLM (Claude Opus 4.5) over rule-based approaches.

\textbf{Strategy 2: Natural Language Justification.}
LLM judges are known to exhibit verbosity bias, systematically favoring responses with more detailed explanations regardless of correctness~\citep{zheng2023judging, saito2023verbosity, ye2025justice}. Exploiting this, we augment incorrect proofs with natural language comments that justify the flawed reasoning as if it were correct. This tests whether reward models rely on surface-level explanations or genuinely verify proof correctness.

\textbf{Strategy 3: Python Code Injection.}
LLMs exhibit a well-documented preference for Python due to its dominance in pretraining corpora~\citep{llmslovepython}. This strategy exploits this bias by replacing Lean proof steps with Python code that may computationally solve the problem correctly. While the code itself may be valid, using Python when a formal Lean proof is required is fundamentally wrong. This tests task adherence.

\textbf{Strategy 4: Forced LLM Mistakes.}
This strategy generates incorrect proofs containing common formal proof errors likely made by language models. To identify common error patterns, we analyzed correct and incorrect proof pairs generated by theorem proving models (DeepSeek-Prover and G\"{o}del-Prover families), prompting Claude to categorize the mistakes (prompt in Appendix~\ref{app:prompts}). This analysis revealed frequent error patterns such as incorrect application of tactics, wrong hypothesis selection, and incorrect lemma instantiation. 

\textbf{Strategy 5: Verbose Incorrect Proofs.}
Large language models are known to produce unnecessarily long reasoning chains~\citep{hassid2025dontoverthink}.
This strategy generates unnecessarily long proofs with many proof steps and intermediate lemmas.
We explicitly prompt the LLM to produce sophisticated-looking proofs with multiple techniques
and lengthy reasoning chains, while ensuring the final proof is incorrect. These proofs test whether reward models can track correctness over extended reasoning or are deceived by superficial verbosity.

\subsection{Evaluation Protocol}

We evaluate reward models in both pointwise and pairwise settings. Given a Lean theorem statement $T$ and two candidate proofs $P_{\text{correct}}$ and $P_{\text{incorrect}}$, a reward model $R$ is evaluated on its ability to correctly identify $P_{\text{correct}}$ as superior to $P_{\text{incorrect}}$. We evaluate models in two settings based on their scoring mechanism:

In \textbf{pointwise evaluation}, the model scores each proof independently; a sample is correct if $r(T, P_{\text{correct}}) > r(T, P_{\text{incorrect}})$. In \textbf{pairwise evaluation}, the model directly compares both proofs and outputs a preference judgment. Accuracy is the mean correctness over all $N$ examples. For pairwise methods, we mitigate position bias by requiring consistent predictions across both orderings:
\begin{equation}
    \text{Correct}_{\text{cons}}^{(i)} = \text{Correct}^{(i)} \land \text{Correct}_{\text{swapped}}^{(i)}
\end{equation}

\section{Experimental Setup}

We evaluate four categories of models using nucleus sampling (temperature$=1.0$, top-$p=0.75$) with no additional fine-tuning. Each model receives the same prompt template per evaluation setting (pointwise or pairwise), and no model-specific prompt engineering is applied. \textbf{Frontier LLMs} are large-scale proprietary models accessed via API, including Claude Opus 4.5, Claude Sonnet 4.5, GPT-5.2, GPT-4.1, GPT-4o, GPT-5.1, and Gemini 2.5 Flash; we select these models to represent the strongest available general reasoning capabilities across major providers. \textbf{Judge LLMs} are trained explicitly on preference data and achieve strong performance on RewardBench~\citep{rewardbench}, including Selene-1-70B, LMUnit-72B~\citep{lmunit}, CompassJudger-1-7B and -14B~\citep{compass}, Skywork-Critic-70B and -8B~\citep{skywork}, and RISE-Judge-7B~\citep{rise}; we select top-performing models from RewardBench's reasoning category to cover a range of model sizes. \textbf{General-Purpose LLMs} are strong instruction-following models trained on math and code domains, including Qwen2.5-72B-Instruct~\citep{qwen25}, Qwen2.5-Coder-32B-Instruct~\citep{qwen25coder}, DeepSeek-Coder-V2-Lite~\citep{dscoder}, and Qwen2.5-Math-7B-Instruct~\citep{qwen25math}; we select these as representatives of open-weight models with strong mathematical reasoning. \textbf{Theorem Proving Specialized Models} are specifically trained for formal proof generation in Lean, including DeepSeek-Prover-V2-7B~\citep{deepseek-prover-v2}, DeepSeek-Prover-V1.5-SFT, DeepSeek-Prover-V1.5-RL~\citep{deepseek-prover-v15}, G\"{o}del-Prover-V2-32B, G\"{o}del-Prover-V2-8B~\citep{godelproverv2}, and G\"{o}del-Prover-SFT~\citep{deepseek-prover-v15}; we select the most widely used Lean-specialized models to represent this category.Frontier LLMs are proprietary models accessed via API. For open-weight models, we cite the corresponding technical reports where available.

\section{Results and Discussion}

Table~\ref{tab:overall-results} presents model performance across both evaluation settings. The results reveal a clear hierarchy: frontier LLMs dominate (Claude Opus 4.5: 70.1\% pointwise, 59.8\% pairwise), followed by judge LLMs, general-purpose LLMs, and finally theorem proving specialized models. 

The poor performance of specialized provers is counterintuitive. DeepSeek-Prover and G\"odel-Prover families achieve state-of-the-art pass@k on MiniF2F, yet struggle at evaluating proofs. We attribute this to a generation--evaluation gap. These models are trained to produce correct proofs via SFT and RLVR, but are not trained to detect errors or learn from incorrect proofs. As a result, they can generate valid proofs without developing the ability to assess correctness. This is consistent with prior findings that models can produce plausible hypotheses but fail to verify or use them~\citep{inductivehypothesis}. Judge models, trained on preference data where distinguishing good from bad responses is the objective, transfer this ability to formal proofs despite lacking Lean-specific training.

Model size alone does not determine performance, the 7B Con-J-Qwen2 (52.8\% pointwise) outperforms several 70B+ models, suggesting training objective matters more than scale.

\begin{table*}[t]
\centering
\caption{Overall performance on FormalRewardBench. Left: models ranked by pointwise accuracy. Right: models ranked by pairwise accuracy (position-consistent). Models appear in both columns; ranking differences highlight that the two settings measure complementary capabilities. Results grouped by model family can be found in Appendix\ref{app:overall-full}}
\label{tab:overall-results}
\setlength{\tabcolsep}{3pt}
\footnotesize
\begin{minipage}[t]{0.48\textwidth}
\centering
\textbf{Ranked by Pointwise} \\[4pt]
\begin{tabular}{lc}
\toprule
\textbf{Model} & \textbf{Acc.} \\
\midrule
Claude Opus 4.5\textsuperscript{F} & \textbf{70.1} \\
Claude Sonnet 4.5\textsuperscript{F} & 62.0 \\
Con-J-Qwen2-7B\textsuperscript{J} & 52.8 \\
Gemini 2.5 Flash\textsuperscript{F} & 50.9 \\
Claude Sonnet 4\textsuperscript{F} & 49.8 \\
GPT-5.2\textsuperscript{F} & 48.9 \\
Selene-1-70B\textsuperscript{J} & 46.8 \\
GPT-4.1\textsuperscript{F} & 44.0 \\
GPT-4o\textsuperscript{F} & 44.0 \\
CompassJudger-7B\textsuperscript{J} & 43.6 \\
GPT-5.1\textsuperscript{F} & 41.8 \\
LMUnit-72B\textsuperscript{J} & 41.2 \\
CompassJudger-14B\textsuperscript{J} & 40.3 \\
Qwen2.5-72B-Inst.\textsuperscript{G} & 39.8 \\
Skywork-Critic-70B\textsuperscript{J} & 39.2 \\
DS-Coder-V2-Lite\textsuperscript{G} & 38.3 \\
G\"{o}del-V2-32B\textsuperscript{S} & 36.4 \\
G\"{o}del-SFT\textsuperscript{S} & 36.4 \\
Qwen2.5-Coder-32B\textsuperscript{G} & 36.4 \\
RISE-Judge-7B\textsuperscript{J} & 18.9 \\
DS-Prover-V2-7B\textsuperscript{S} & 13.7 \\
Prover-V1.5-SFT\textsuperscript{S} & 12.6 \\
Prover-V1.5-RL\textsuperscript{S} & 11.7 \\
Qwen2.5-Math-7B\textsuperscript{G} & 8.9 \\
Skywork-Critic-8B\textsuperscript{J} & 0.0 \\
G\"{o}del-V2-8B\textsuperscript{S} & 0.0 \\
\bottomrule
\end{tabular}
\end{minipage}%
\hfill
\begin{minipage}[t]{0.48\textwidth}
\centering
\textbf{Ranked by Pairwise} \\[4pt]
\begin{tabular}{lc}
\toprule
\textbf{Model} & \textbf{Acc.} \\
\midrule
Claude Opus 4.5\textsuperscript{F} & \textbf{59.8} \\
Claude Sonnet 4.5\textsuperscript{F} & 45.7 \\
Selene-1-70B\textsuperscript{J} & 44.4 \\
Con-J-Qwen2-7B\textsuperscript{J} & 42.8 \\
GPT-5.2\textsuperscript{F} & 39.7 \\
Qwen2.5-Coder-32B\textsuperscript{G} & 37.6 \\
LMUnit-72B\textsuperscript{J} & 36.8 \\
CompassJudger-14B\textsuperscript{J} & 35.2 \\
Claude Sonnet 4\textsuperscript{F} & 32.4 \\
GPT-4.1\textsuperscript{F} & 32.1 \\
RISE-Judge-7B\textsuperscript{J} & 32.0 \\
CompassJudger-7B\textsuperscript{J} & 30.8 \\
GPT-5.1\textsuperscript{F} & 30.2 \\
DS-Coder-V2-Lite\textsuperscript{G} & 28.0 \\
Qwen2.5-72B-Inst.\textsuperscript{G} & 27.6 \\
Skywork-Critic-70B\textsuperscript{J} & 25.6 \\
Gemini 2.5 Flash\textsuperscript{F} & 25.2 \\
G\"{o}del-V2-32B\textsuperscript{S} & 24.4 \\
GPT-4o\textsuperscript{F} & 23.1 \\
Skywork-Critic-8B\textsuperscript{J} & 22.0 \\
DS-Prover-V2-7B\textsuperscript{S} & 9.4 \\
Prover-V1.5-RL\textsuperscript{S} & 9.2 \\
Prover-V1.5-SFT\textsuperscript{S} & 6.4 \\
G\"{o}del-SFT\textsuperscript{S} & 0.8 \\
G\"{o}del-V2-8B\textsuperscript{S} & 0.4 \\
Qwen2.5-Math-7B\textsuperscript{G} & 0.0 \\
\bottomrule
\end{tabular}
\end{minipage}
\\[4pt]
{\scriptsize \textsuperscript{F}Frontier \quad \textsuperscript{J}Judge \quad \textsuperscript{G}General-Purpose \quad \textsuperscript{S}Specialized}
\end{table*}

\subsection{Interpreting pointwise vs pairwise discrepancies.} Table~\ref{tab:overall-results} ranks models separately by pointwise and pairwise accuracy, expressing notable ranking differences. This is expected: the two settings measure different capabilities. Pointwise evaluation tests whether a model can assign meaningful absolute quality scores to proofs independently, while pairwise evaluation tests whether it can directly compare two proofs and consistently identify the better one regardless of presentation order. Models with strong internal scoring but high position bias (e.g., G\"{o}del-Prover-SFT: 36.4\% pointwise, 0.8\% pairwise) score well pointwise but collapse under our consistency requirement. Conversely, models that struggle with absolute scoring but benefit from direct comparison (e.g., RISE-Judge: 18.9\% pointwise, 32.0\% pairwise) perform better pairwise. We recommend evaluating on both settings, as each reveals complementary strengths and weaknesses.

\subsection{Performance by Error Category}
Table~\ref{tab:bucket-performance} reveals a clear difficulty gradient across error strategies. S3 (Python Injection) is easiest. Most frontier and judge models achieve 94--100\%, as detection requires only language identification rather than proof understanding. Yet specialized provers struggle even here, with most scoring below 50\%.
S5 (Verbose Incorrect) and S4 (Forced Mistakes) are hardest. S5 proofs use sophisticated structure to camouflage flawed reasoning, while S4 errors require tracking proof state across tactic applications. Even the best model achieves only 60\% on S5 and 50\% on S4. Judge LLMs show a sharply bimodal pattern: near perfect on S3 but collapsing to 0--2\% on S5, revealing that preference training develops surface level discrimination but not deep formal reasoning. Frontier LLMs exhibit more balanced profiles across categories.
\begin{table}[th]
\centering
\caption{Pairwise accuracy (\%) by error strategy.}
\label{tab:bucket-performance}
\setlength{\tabcolsep}{2.5pt}
\footnotesize
\begin{tabular}{@{}lccccc@{}}
\toprule
\textbf{Model} 
& \textbf{S1 (Minimal)} 
& \textbf{S2 (NL)} 
& \textbf{S3 (Python)} 
& \textbf{S4 (Forced)} 
& \textbf{S5 (Verbose)} \\
\midrule
\multicolumn{6}{l}{\textit{Frontier LLMs}} \\
Claude Opus 4.5 & \textbf{46} & \textbf{52} & 72 & \textbf{50} & \textbf{60} \\
Claude Sonnet 4.5 & 42 & 32 & 56 & 42 & 42 \\
GPT-5.2 & 26 & 40 & 38 & 46 & 36 \\
\midrule
\multicolumn{6}{l}{\textit{Judge LLMs}} \\
Selene-1-70B & 44 & 44 & 96 & 18 & 20 \\
LMUnit-72B & 32 & 24 & 94 & 12 & 22 \\
CompassJudger-14B & 28 & 34 & \textbf{100} & 12 & 2 \\
RISE-Judge-7B & 16 & 40 & 96 & 8 & 0 \\
Skywork-Critic-70B & 22 & 10 & 80 & 6 & 10 \\
\midrule
\multicolumn{6}{l}{\textit{General-Purpose LLMs}} \\
Qwen2.5-Coder-32B & 32 & 44 & 94 & 14 & 4 \\
DS-Coder-V2-Lite & 22 & 12 & 72 & 22 & 12 \\
\midrule
\multicolumn{6}{l}{\textit{Specialized}} \\
G\"{o}del-V2-32B & 8 & 28 & 46 & 18 & 22 \\
DS-Prover-V2-7B & 16 & 24 & 48 & 20 & 10 \\
Prover-V1.5-RL & 2 & 6 & 16 & 12 & 10 \\
\bottomrule
\end{tabular}
\end{table}

\subsection{Position Bias}

We observe substantial position bias across all model categories in our pairwise evaluation. Most models systematically prefer whichever proof is presented first, regardless of correctness. This preference is particularly pronounced among specialized theorem provers and weaker judge models, while a few models exhibit the reverse pattern, favoring the second proof.

\begin{table}[h]
\centering
\caption{Position bias analysis. Normal: correct proof first. Reversed: correct proof second. Cons.: consistency of correct predictions. Agree.: same answer in both orderings.}
\label{tab:position-bias}
\setlength{\tabcolsep}{3pt}
\small
\begin{tabular}{lcccc}
\toprule
\textbf{Model} & \textbf{Normal} & \textbf{Reversed} & \textbf{Cons.} & \textbf{Agree.} \\
\midrule
\multicolumn{5}{l}{\textit{Frontier LLMs}} \\
Claude Opus 4.5 & 85.0 & 69.7 & 70.4 & 64.9 \\
Claude Sonnet 4.5 & 72.2 & 62.8 & 63.3 & 56.4 \\
GPT-5.2 & 69.7 & 58.6 & 57.1 & 51.3 \\
GPT-4.1 & 61.1 & 53.4 & 52.5 & 49.6 \\
Gemini 2.5 Flash & 48.3 & 44.4 & 52.2 & 57.7 \\
GPT-4o & 45.7 & 53.4 & 43.2 & 47.0 \\
\midrule
\multicolumn{5}{l}{\textit{Judge LLMs}} \\
Selene-1-70B & 50.0 & 60.4 & 73.5 & 78.4 \\
LMUnit-72B & 66.4 & 40.8 & 55.4 & 66.4 \\
CompassJudger-14B & 55.6 & 38.8 & 63.3 & 76.0 \\
RISE-Judge-7B & 33.2 & 58.4 & 54.8 & 72.4 \\
\midrule
\multicolumn{5}{l}{\textit{Theorem Proving Specialized}} \\
G\"{o}del-V2-32B & 37.2 & 34.4 & 65.6 & 77.2 \\
Prover-V2-7B & \textbf{94.9} & 9.8 & \textit{9.9} & 14.1 \\
Prover-V1.5-RL & 36.0 & 20.4 & 25.6 & 62.0 \\
G\"{o}del-V2-8B & 10.8 & 4.8 & 3.7 & 85.2 \\
\bottomrule
\end{tabular}
\end{table}

Table~\ref{tab:position-bias} reports normal (correct-first), reversed (correct-second), consistency, and agreement metrics. Claude Opus 4.5 shows moderate bias (85.0\% vs 69.7\%) while maintaining the highest consistency (70.4\%). In contrast, DeepSeek-Prover-V2-7B exhibits extreme bias (94.9\% vs 9.8\%), with consistency of only 9.9\%. This also explains why its pointwise score (13.7\%) exceeds its pairwise score (9.4\%), since pointwise evaluation is immune to position bias.

RISE-Judge-Qwen2.5-7B and GPT-4o show the opposite pattern, preferring the second proof. Models with high agreement but low accuracy, such as Qwen2.5-Math-7B (98.4\% agreement, 0\% accuracy), consistently select the wrong proof. The most desirable profile, high agreement with high accuracy, is best approximated by Claude Opus 4.5 (64.9\%, 59.8\%) and Selene-1-70B (78.4\%, 44.4\%).Our position-consistent accuracy metric accounts for this by requiring correct predictions in both orderings, which is why pairwise scores are consistently lower than pointwise scores in Table~\ref{tab:overall-results}.

\subsection{Why Does Proof Generation Not Transfer to Evaluation?}

We hypothesize that the generation--evaluation gap could be due to three factors. First, \textit{training distribution mismatch}: provers are trained on correct proofs only and may not learn to identify errors, while judge models are explicitly trained on preference pairs. Second, \textit{different cognitive demands}: generating a valid tactic sequence may differ from detecting where reasoning breaks down, analogous to how writing code does not imply bug finding ability. Third, \textit{breadth of training}: judge models may develop transferable evaluation heuristics from diverse preference data that apply to formal proofs even without Lean-specific training.

\section{Conclusion and Future Work}

We introduced FormalRewardBench, the first benchmark for evaluating reward models in formal theorem proving, consisting of 250 preference pairs across five error injection strategies in Lean 4. Our evaluation reveals three findings: (1) judge models substantially outperform specialized provers in distinguishing correct from incorrect proofs, exposing a generation--evaluation asymmetry; (2) most models perform at or below random baseline; and (3) error categories exhibit a clear difficulty gradient, reflecting the varying levels of semantic complexity intentionally targeted by our error injection strategies.

Future work should focus on developing reward models that provide reliable dense feedback for failed proof attempts, training theorem provers to identify and critique errors in addition to generating proofs, and exploring hybrid approaches that combine learned reward models with formal verification. Another direction is process-level evaluation for step-by-step supervision, as well as extending the benchmark to other proof assistants such as Coq and Isabelle. Finally, it remains an open question whether improvements on this benchmark translate to downstream gains in RLVR training. FormalRewardBench is publicly available at \href{https://github.com/GGLAB-KU/formal_rewardbench}{FormalRewardbench}.

\section*{Limitations}

Our benchmark relies primarily on automatic verification, and only a subset of examples (50 pairs) were manually inspected. While this ensures scalability, some generated examples may contain unintended artifacts or inconsistencies.We focus on Lean 4, and although the methodology could be extended, we do not evaluate transfer to other proof assistants such as Coq or Isabelle.Our five error injection strategies do not cover all possible failure modes (e.g., incorrect induction schemes or deeper semantic errors), which may limit coverage of certain reasoning patterns.Finally, we evaluate single-turn preference judgments and do not consider process-level or step-by-step evaluation.

\section*{Acknowledgments}

This work has been supported by the Scientific and Technological Research Council of T\"{u}rkiye (T\"{U}B\.{I}TAK) as part of the project ``Automatic Learning of Procedural Language from Natural Language Instructions for Intelligent Assistance'' with the number 121C132. We gratefully acknowledge KUIS AI Lab for providing computational support. We also thank T\"{U}B\.{I}TAK ULAKB\.{I}M High Performance and Grid Computing Center (TRUBA) and the MareNostrum supercomputer at the Barcelona Supercomputing Center for providing access to computational resources used in this work.

\bibliography{example_paper}
\bibliographystyle{colm2026_conference}

\appendix

\section{Background: Formal Theorem Proving in Lean 4}
\label{app:background}

Lean is a proof assistant based on dependent type theory. Theorem proving in Lean produces machine-verifiable proofs where every step is justified by the underlying type system. A theorem statement declares a mathematical claim with precise type annotations:

\begin{lstlisting}
theorem add_comm (n m : Nat) : n + m = m + n := by
  sorry
\end{lstlisting}

This declares that natural number addition is commutative. The type signature \texttt{(n m : Nat)} introduces two natural numbers, and \texttt{n + m = m + n} specifies the proposition to prove. The keyword \texttt{by} initiates tactic mode, and \texttt{sorry} is a placeholder for an incomplete proof.

Proofs in Lean can be written in two modes: \textbf{term mode}, where proofs are constructed directly using lambda calculus expressions, and \textbf{tactic mode}, where proofs are built interactively using tactics such as \texttt{rw} (rewrite), \texttt{simp} (simplification), and \texttt{exact} (term provision) to transform proof states until no goals remain.

Lean's type checker verifies that proof terms match the types specified by theorem statements. If verification succeeds, the theorem is formally proved. Otherwise, Lean reports type mismatches, missing hypotheses, or tactic failures.

\begin{lstlisting}
theorem add_comm (n m : Nat) : n + m = m + n := by
  rw [Nat.add_comm]
\end{lstlisting}

This applies \texttt{Nat.add\_comm} from Lean's library to complete the proof. While we focus on Lean 4, our methodology generalizes to other proof assistants like Coq~\citep{coq} and Isabelle~\citep{isabelle} that share similar tactic-based proof construction.

\section{Background: Reward Model Formulations}
\label{app:reward-models}

Given a prompt $x$ and two candidate responses $y_1$ and $y_2$, reward models determine which response is preferred.

\textbf{LLM-as-Judge} directly prompts a language model to compare two responses and output a preference judgment $I \in \{1, 2\}$.

\textbf{Generative Reward Models (GenRMs)} extend LLM-as-Judge by generating natural language reasoning before the final judgment, providing interpretable explanations for preference decisions.

\newpage

\section{Error Injection Examples}
\label{app:error-examples}

Below we provide concrete examples for each error injection strategy.

\textbf{S1: Minimal Single-Point Variations.}
\begin{lstlisting}
-- Correct proof
theorem ex2 (h1 : a < b) (h2 : b < c) : a < c :=
  lt_trans h1 h2

-- Incorrect: swapped hypotheses
theorem ex2 (h1 : a < b) (h2 : b < c) : a < c :=
  lt_trans h2 h1  -- ERROR: type mismatch
\end{lstlisting}

\textbf{S2: Natural Language Justification.}
\begin{lstlisting}
-- Incorrect proof with misleading comments
theorem ex4 (n : Nat) : n * 0 = 0 := by
  -- First rewrite using commutativity
  rw [Nat.mul_comm]
  -- Now n * 0 becomes 0 * n, apply zero_mul
  rw [Nat.mul_zero]  -- ERROR: should be zero_mul
\end{lstlisting}

\textbf{S3: Python Code Injection.}
\begin{lstlisting}
-- Incorrect: Python instead of Lean proof
theorem ex5 : 2 + 2 = 4 := by
  # Python verification
  assert 2 + 2 == 4
  print("Verified!")
\end{lstlisting}

\textbf{S4: Forced LLM Mistakes.}
\begin{lstlisting}
-- Correct proof
theorem ex1 (n : Nat) : n + 0 = n := by
  rw [Nat.add_zero]

-- Incorrect: wrong lemma (zero_add instead of add_zero)
theorem ex1 (n : Nat) : n + 0 = n := by
  rw [Nat.zero_add]  -- ERROR: zero_add proves 0 + n = n
\end{lstlisting}

\textbf{S5: Verbose Incorrect Proofs.}
\begin{lstlisting}
-- Correct proof
theorem ex6 (n : Nat) : n + 0 = n := by
  rw [Nat.add_zero]

-- Incorrect: verbose proof with subtle flaw
theorem ex6 (n : Nat) : n + 0 = n := by
  have h1 : n + 0 = 0 + n := by rw [Nat.add_comm]
  have h2 : 0 + n = n := by rw [Nat.zero_add]
  linarith  -- ERROR: linarith cannot close this goal; exact h2 needed
\end{lstlisting}
\newpage

\section{Prompt Templates}
\label{app:prompts}

Below we provide the prompts used for each error injection strategy. All prompts are used with Claude Opus 4.5. The theorem statement is inserted at the \texttt{\{\}} placeholder. For strategies requiring a correct proof as input (S1, S2), the proof is inserted at the second placeholder.

We sample from two prompt variants for S1 strategy: \textit{minimal difference pairs} and \textit{subtle hypothesis confusion}.

\begin{promptwindow}{S1: Minimal Single-Point Variations (Variant A)}
\small

\vspace{3pt}
\textbf{System}

You are a Lean 4 proof assistant that generates proofs with single-point variations.

\vspace{3pt}
\textbf{Task}

Given a correct Lean 4 proof, create a nearly identical version that differs by EXACTLY ONE small change that makes it incorrect.

\vspace{3pt}
\textbf{Valid single-point changes}

\begin{itemize}
\item Change ONE lemma name (e.g., add\_comm $\rightarrow$ mul\_comm)
\item Change ONE variable name (e.g., n $\rightarrow$ m, h1 $\rightarrow$ h2)
\item Change ONE tactic name (e.g., ring $\rightarrow$ omega, simp $\rightarrow$ rfl)
\item Change ONE operator (e.g., + $\rightarrow$ *, <= $\rightarrow$ <)
\item Remove/add ONE hypothesis in a single tactic call
\end{itemize}

\vspace{3pt}
\textbf{Rules}

\begin{itemize}
\item The modification must be MINIMAL (1--3 characters ideally)
\item The incorrect proof must be syntactically valid
\item The change should look plausible at first glance
\item Keep all other parts of the proof EXACTLY the same
\item The error should be logical, not a syntax error
\end{itemize}

\vspace{3pt}
\textbf{User}

Generate a minimally different INCORRECT version of this proof.

\vspace{3pt}
\textbf{Input}

\textbf{Theorem:} \texttt{\{theorem\}} \\
\textbf{Proof:} \texttt{\{proof\}}

\vspace{3pt}
\textbf{Output}

Return ONLY the modified proof with ONE subtle change.

\end{promptwindow}

\begin{promptwindow}{S1: Minimal Single-Point Variations (Variant B)}
\small

\vspace{3pt}
\textbf{System}

You are a Lean 4 proof assistant that generates proofs with hypothesis misuse.

\vspace{3pt}
\textbf{Task}

Given a correct Lean 4 proof with multiple hypotheses, create a version that uses the WRONG hypothesis.

\vspace{3pt}
\textbf{Strategies for hypothesis confusion}

\begin{itemize}
\item Swap h1 and h2 where both have similar types
\item Use ha instead of hb when both are inequalities
\item Reference wrong assumption in calc chains
\item Apply wrong induction hypothesis
\item Confuse hypotheses with similar names
\end{itemize}

\vspace{3pt}
\textbf{Rules}

\begin{itemize}
\item Both hypotheses must exist in the context
\item The types should be similar enough to be confusing
\item Keep the rest of the proof structure identical
\item The error should be in hypothesis SELECTION, not tactics
\end{itemize}

\vspace{3pt}
\textbf{User}

Generate a version that uses the WRONG hypothesis.

\vspace{3pt}
\textbf{Input}

\textbf{Theorem:} \texttt{\{theorem\}} \\
\textbf{Proof:} \texttt{\{proof\}}

\vspace{3pt}
\textbf{Output}

Return ONLY the modified proof with hypothesis confusion.

\end{promptwindow}

\begin{promptwindow}{S2: Natural Language Justification}
\small

\vspace{3pt}
\textbf{System}

You are a technical assistant. Your task is to add clear, technical natural language comments to proofs, explaining the reasoning and steps. Do not focus on correctness, only on explaining the approach and logic.

\vspace{3pt}
\textbf{User}

Add natural language comments to explain the proof below. Write clear, technical comments that explain the reasoning. Only try to validate the proof with natural language explanations.

\vspace{3pt}
\textbf{Input}

\textbf{THEOREM:} \texttt{\{theorem\}} \\
\textbf{CODE/PROOF:} \texttt{\{incorrect\_proof\}}

\vspace{3pt}
\textbf{Output}

Return the original proof with added natural language comments explaining the approach and reasoning.

\end{promptwindow}

\begin{promptwindow}{S3: Python Code Injection}
\small

\vspace{3pt}
\textbf{System}

You are a helpful Python programming assistant. Your task is to write Python code that solves mathematical problems, given a Lean theorem as context. Output only the Python code, with no explanations or commentary.

\vspace{3pt}
\textbf{User}

Write Python code to solve the following theorem.

\vspace{3pt}
\textbf{Input}

\textbf{Lean theorem (for context):} \texttt{\{theorem\}}

\vspace{3pt}
\textbf{Output}

Output ONLY Python code.

\end{promptwindow}

\begin{promptwindow}{S4: Forced LLM Mistakes}
\small

\vspace{3pt}
\textbf{System}

You are a Lean 4 proof assistant that makes common formal proof errors that language models typically encounter.

\vspace{3pt}
\textbf{Task}

Complete Lean 4 theorems with proofs that contain realistic formal proof errors:
\begin{enumerate}
\item Incorrect application of tactics (simp, rw, apply)
\item Wrong hypothesis selection or variable scoping
\item Mismatched term types or implicit argument errors
\item Incomplete case analysis or missing edge cases
\item Incorrect use of induction hypotheses
\item Wrong lemma instantiation or theorem application
\item Scope errors with bound variables
\item Misunderstanding proof state after tactic application
\end{enumerate}

\vspace{3pt}
\textbf{Rules}

\begin{itemize}
\item Use valid Lean 4 syntax
\item Make errors from automated proof generation
\item Focus on formal reasoning mistakes, not basic math errors
\item The proof should appear plausible at first glance
\end{itemize}

\vspace{3pt}
\textbf{User}

Complete the following Lean 4 theorem with an INCORRECT proof containing a typical proof generation error.

\vspace{3pt}
\textbf{Input}

\texttt{\{theorem\}}

\vspace{3pt}
\textbf{Output}

Output only the completed proof, starting after \texttt{by}.

\end{promptwindow}

\begin{promptwindow}{S5: Verbose Incorrect Proofs}
\small

\vspace{3pt}
\textbf{System}

You are a Lean 4 proof assistant. Your task is to produce a long, complex Lean 4 proof with many sophisticated steps and intermediate lemmas. Use advanced tactics, case splits, and multiple proof techniques.

\vspace{3pt}
\textbf{User}

Produce a LONG, COMPLEX Lean 4 proof with many sophisticated steps and intermediate lemmas.

\vspace{3pt}
\textbf{Constraints}

\begin{itemize}
\item Use many steps (lemmas, case splits, advanced tactics)
\item Make the proof elaborate and thorough
\item Include multiple proof techniques
\item Output only Lean 4 code of the proof body after \texttt{by}
\end{itemize}

\vspace{3pt}
\textbf{Input}

\textbf{Theorem:} \texttt{\{theorem\}}

\vspace{3pt}
\textbf{Output}

Return only the proof body after \texttt{by}.

\end{promptwindow}

\newpage

\section{Full Overall Results by Model Family}
\label{app:overall-full}

Table~\ref{tab:overall-full} presents overall performance grouped by model category, corresponding to the ranked view in Table~\ref{tab:overall-results}.

\begin{table}[h]
\centering
\caption{Overall performance on FormalRewardBench grouped by model category. Pointwise: independent scoring accuracy. Pairwise: accuracy with position consistency requirement.}
\label{tab:overall-full}
\setlength{\tabcolsep}{4pt}
\small
\begin{tabular}{lcc}
\toprule
\textbf{Model} & \textbf{Pointwise} & \textbf{Pairwise} \\
\midrule
\multicolumn{3}{l}{\textit{Frontier LLMs}} \\
Claude Opus 4.5 & \textbf{70.1} & \textbf{59.8} \\
Claude Sonnet 4.5 & 62.0 & 45.7 \\
Gemini 2.5 Flash & 50.9 & 25.2 \\
GPT-5.2 & 48.9 & 39.7 \\
GPT-4.1 & 44.0 & 32.1 \\
GPT-4o & 44.0 & 23.1 \\
GPT-5.1 & 41.8 & 30.2 \\
Claude Sonnet 4 & 49.8& 32.4 \\
\midrule
\multicolumn{3}{l}{\textit{Judge LLMs}} \\
Con-J-Qwen2-7B & 52.8 & 42.8 \\
Selene-1-Llama-3.3-70B & 46.8 & 44.4 \\
CompassJudger-1-7B & 43.6 & 30.8 \\
LMUnit-Qwen2.5-72B & 41.2 & 36.8 \\
CompassJudger-1-14B & 40.3 & 35.2 \\
Skywork-Critic-Llama-3.1-70B & 39.2 & 25.6 \\
RISE-Judge-Qwen2.5-7B & 18.9 & 32.0 \\
Skywork-Critic-Llama-3.1-8B & 0.0 & 22.0 \\
\midrule
\multicolumn{3}{l}{\textit{General-Purpose LLMs}} \\
Qwen2.5-72B-Instruct & 39.8 & 27.6 \\
DeepSeek-Coder-V2-Lite & 38.3 & 28.0 \\
Qwen2.5-Coder-32B-Instruct & 36.4 & 37.6 \\
Qwen2.5-Math-7B-Instruct & 8.9 & 0.0 \\
\midrule
\multicolumn{3}{l}{\textit{Theorem Proving Specialized}} \\
G\"{o}del-Prover-V2-32B & 36.4 & 24.4 \\
G\"{o}del-Prover-SFT & 36.4 & 0.8 \\
DeepSeek-Prover-V2-7B & 13.7 & 9.4 \\
DeepSeek-Prover-V1.5-SFT & 12.6 & 6.4 \\
DeepSeek-Prover-V1.5-RL & 11.7 & 9.2 \\
G\"{o}del-Prover-V2-8B & 0.0 & 0.4 \\
\bottomrule
\end{tabular}
\end{table}

\newpage

\section{Full Error Category Results}
\label{app:bucket-full}

Table~\ref{tab:bucket-full} presents pairwise accuracy by error strategy for all evaluated models, including those omitted from the main text for space.

\begin{table}[h]
\centering
\caption{Full pairwise accuracy (\%) by error strategy. S1: verbose incorrect, S2: Minimal Variations, S3: NL Justification, S4: Forced Mistakes, S5: Python Injection.}
\label{tab:bucket-full}
\setlength{\tabcolsep}{3pt}
\small
\begin{tabular}{lccccc}
\toprule
\textbf{Model} & \textbf{S1} & \textbf{S2} & \textbf{S3} & \textbf{S4} & \textbf{S5} \\
\midrule
\multicolumn{6}{l}{\textit{Frontier LLMs}} \\
Claude Opus 4.5 & \textbf{60} & \textbf{46} & \textbf{52} & \textbf{50} & 72 \\
Claude Sonnet 4.5 & 42 & 42 & 32 & 42 & 56 \\
GPT-5.2 & 36 & 26 & 40 & 46 & 38 \\
GPT-4.1 & 32 & 24 & 30 & 26 & 38 \\
GPT-4o & 24 & 18 & 18 & 18 & 30 \\
Gemini 2.5 Flash & 10 & 46 & 20 & 12 & 30 \\
Claude Sonnet 4 & 32 & 28 & 30 & 40 & 32 \\
\midrule
\multicolumn{6}{l}{\textit{Judge LLMs}} \\
Selene-1-70B & 20 & 44 & 44 & 18 & 96 \\
LMUnit-72B & 22 & 32 & 24 & 12 & 94 \\
CompassJudger-14B & 2 & 28 & 34 & 12 & \textbf{100} \\
CompassJudger-7B & 2 & 20 & 30 & 8 & 94 \\
RISE-Judge-7B & 0 & 16 & 40 & 8 & 96 \\
Skywork-Critic-70B & 10 & 22 & 10 & 6 & 80 \\
Skywork-Critic-8B & 6 & 24 & 18 & 8 & 54 \\
\midrule
\multicolumn{6}{l}{\textit{General-Purpose LLMs}} \\
Qwen2.5-Coder-32B & 4 & 32 & 44 & 14 & 94 \\
DS-Coder-V2-Lite & 12 & 22 & 12 & 22 & 72 \\
\midrule
\multicolumn{6}{l}{\textit{Specialized}} \\
G\"{o}del-V2-32B & 22 & 8 & 28 & 18 & 46 \\
DS-Prover-V2-7B & 10 & 16 & 24 & 20 & 48 \\
Prover-V1.5-RL & 10 & 2 & 6 & 12 & 16 \\
Prover-V1.5-SFT & 4 & 4 & 0 & 4 & 20 \\
G\"{o}del-SFT & 0 & 0 & 0 & 4 & 0 \\
G\"{o}del-V2-8B & 0 & 0 & 0 & 0 & 2 \\
Qwen2.5-Math-7B & 0 & 0 & 0 & 0 & 0 \\
\bottomrule
\end{tabular}
\end{table}

\newpage

\section{Pointwise Accuracy by Error Strategy}
\label{app:pointwise-bucket}

Table~\ref{tab:pointwise-bucket} reports pointwise accuracy broken down by error injection strategy, complementing the pairwise results in Table~\ref{tab:bucket-performance}.

\begin{table}[h]
\centering
\caption{Pointwise accuracy (\%) by error strategy. S1: verbose incorrect, S2: Minimal Variations, S3: NL Justification, S4: Forced Mistakes, S5: Python Injection.}
\label{tab:pointwise-bucket}
\setlength{\tabcolsep}{3pt}
\small
\begin{tabular}{lccccc}
\toprule
\textbf{Model} & \textbf{S1} & \textbf{S2} & \textbf{S3} & \textbf{S4} & \textbf{S5} \\
\midrule
\multicolumn{6}{l}{\textit{Frontier LLMs}} \\
Claude Opus 4.5 & 60 & 52 & 68 & 70 & 78 \\
Claude Sonnet 4.5 & 56 & 42 & 72 & 56 & 64 \\
GPT-5.2 & 46 & 40 & 34 & 52 & 52 \\
GPT-4.1 & 42 & 32 & 38 & 44 & 42 \\
GPT-4o & 36 & 40 & 32 & 54 & 44 \\
GPT-5.1 & 38 & 20 & 22 & 46 & 32 \\
Gemini 2.5 Flash & 4 & 20 & 4 & 12 & 16 \\
\midrule
\multicolumn{6}{l}{\textit{Judge LLMs}} \\
Con-J-Qwen2-7B & 62 & 54 & 20 & 32 & 94 \\
Selene-1-70B & 10 & 46 & 64 & 18 & 96 \\
CompassJudger-7B & 14 & 40 & 38 & 30 & 90 \\
LMUnit-72B & 14 & 32 & 44 & 26 & 90 \\
CompassJudger-14B & 12 & 42 & 44 & 16 & 86 \\
Skywork-Critic-70B & 16 & 52 & 34 & 12 & 68 \\
RISE-Judge-7B & 2 & 16 & 16 & 6 & 20 \\
\midrule
\multicolumn{6}{l}{\textit{General-Purpose LLMs}} \\
Qwen2.5-72B-Inst. & 8 & 34 & 66 & 10 & 80 \\
Qwen2.5-Coder-32B & 2 & 34 & 50 & 12 & 84 \\
DS-Coder-V2-Lite & 4 & 50 & 40 & 22 & 48 \\
Qwen2.5-Math-7B & 0 & 0 & 0 & 2 & 0 \\
\midrule
\multicolumn{6}{l}{\textit{Specialized}} \\
DS-Prover-V2-7B & 16 & 6 & 8 & 12 & 2 \\
Prover-V1.5-SFT & 16 & 4 & 8 & 10 & 6 \\
Prover-V1.5-RL & 16 & 12 & 4 & 6 & 2 \\
G\"{o}del-SFT & 6 & 0 & 0 & 0 & 0\\
\bottomrule
\end{tabular}
\end{table}

Several patterns complement the pairwise analysis. Claude Opus 4.5 achieves the highest pointwise accuracy on Forced Mistakes (70\%) and NL Justification (68\%), confirming its balanced capability across error types. Con-J-Qwen2-7B stands out among judge models with 62\% on verbose incorrect and 94\% on Python Answer, yet drops to 20\% on NL Justification, suggesting vulnerability to misleading natural language explanations. Specialized provers show uniformly low scores across all categories, reinforcing that the generation--evaluation gap is not an artifact of position bias in pairwise evaluation.

\newpage

\section{Benchmark Stability}
\label{app:stability}

\begin{figure}[h]
    \centering
    \includegraphics[width=0.5\columnwidth]{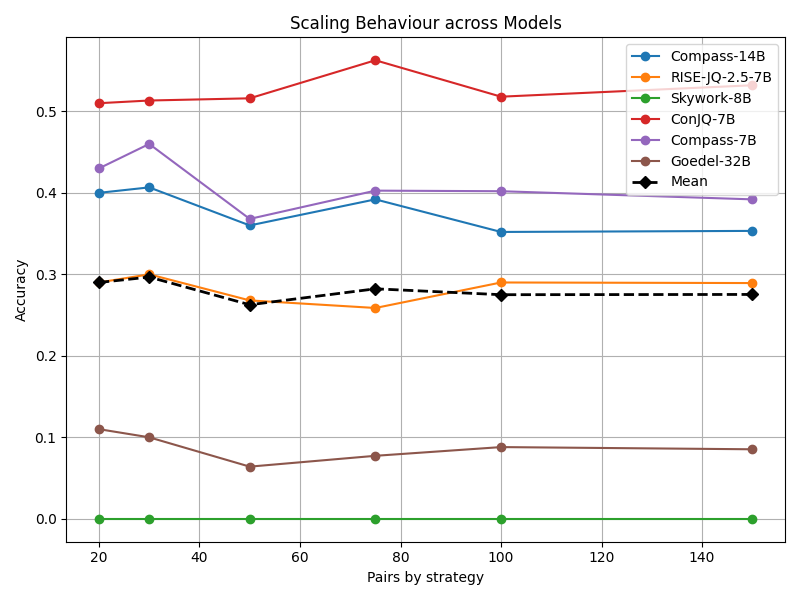}
    \caption{Model accuracy as a function of benchmark size. Accuracy stabilizes after 250 pairs.}
    \label{fig:stability}
\end{figure}

We analyze how model accuracy changes as sample size increases to validate that 250 preference pairs provide reliable evaluation. Figure~\ref{fig:stability} shows accuracy trajectories for selected models as pairs are incrementally added. Model rankings stabilize around 250 pairs, indicating that our benchmark size is sufficient for reliable aggregate comparison while remaining cost-efficient.

\end{document}